%% file: main.tex
\documentclass[10pt,twocolumn,letterpaper]{article}

 \usepackage[pagenumbers]{cvpr} 

\input{preamble}

\definecolor{cvprblue}{rgb}{0.21,0.49,0.74}
\usepackage[pagebackref,breaklinks,colorlinks,citecolor=cvprblue]{hyperref}

\def\paperID{17076} 
\def\confName{CVPR}
\def\confYear{2024}

\title{Input Compression with Positional Consistency for Efficient Training and Inference of Transformer Neural Networks}

\author{Amrit Nagarajan\\
School of ECE\\
Purdue University\\
{\tt\small nagaraj9@purdue.edu}
\and
Anand Raghunathan\\
School of ECE\\
Purdue University\\
{\tt\small raghunathan@purdue.edu}
}

\begin{document}
\maketitle
\input{sections/abstract}    
\input{sections/intro}

\input{sections/icpc_methods}

\input{sections/pos_embed_selection}
\input{sections/icpc_inference}
\input{sections/expts_and_results}

\input{sections/related_work}
\input{sections/conclusion}
\section{Acknowledgement}
This work was supported in part by the Center for the Co-Design of Cognitive Systems (CoCoSys), a JUMP2.0 center sponsored by the Semiconductor Research Corporation (SRC) and DARPA, and in part by SRC under the AIHW program.
{
    \small
    \bibliographystyle{ieeenat_fullname}
    \bibliography{main}
}


\end{document}

%% file: preamble.tex
\usepackage[dvipsnames]{xcolor}

\usepackage[ruled,vlined,linesnumbered]{algorithm2e}
\usepackage{multicol}
\usepackage{enumitem}
\usepackage{multirow}
\usepackage{wrapfig}

%% file: sections/abstract.tex
\begin{abstract}
Transformers have rapidly increased in popularity in recent years, achieving state-of-the-art performance in processing text, images, audio and video. However, Transformers present large computational requirements for both training and inference, and are prone to overfitting during training. To address these challenges, we present Input Compression with Positional Consistency (\textbf{ICPC}), a new data augmentation method that, unlike prior augmentation techniques, simultaneously improves both generalization and training efficiency. ICPC applies varying levels of compression to each training sample in each epoch. This leads to smaller input sequences being processed by the Transformer, and hence faster training, while also alleviating overfitting by presenting each input with different compression levels. We introduce a consistency-aware position selection method in ICPC that enables accurate processing of compressed inputs without any changes to the underlying Transformer architecture. We detail compression-based augmentation methods for four different modalities -- insignificant word pruning for text, resolution modulation for images, spatio-temporal resolution modulation for videos, and spectogram size modulation for audio. ICPC also enables efficient variable-effort inference, where samples are first inferred at high compression levels, and progressively re-evaluated with lower compression for more challenging inputs. On 9 diverse tasks spanning 4 different modalities, ICPC improves accuracy by up to 1\%, while also accelerating training and inference by up to 2.9$\times$ and 2.6$\times$, respectively. Code is available at \url{https://github.com/amrnag/ICPC}.
\end{abstract}

%% file: sections/intro.tex
\vspace{-10pt}
\section{Introduction}
\label{sec:intro}

\begin{figure*}[htb]
\centerline{\includegraphics[width=\linewidth]{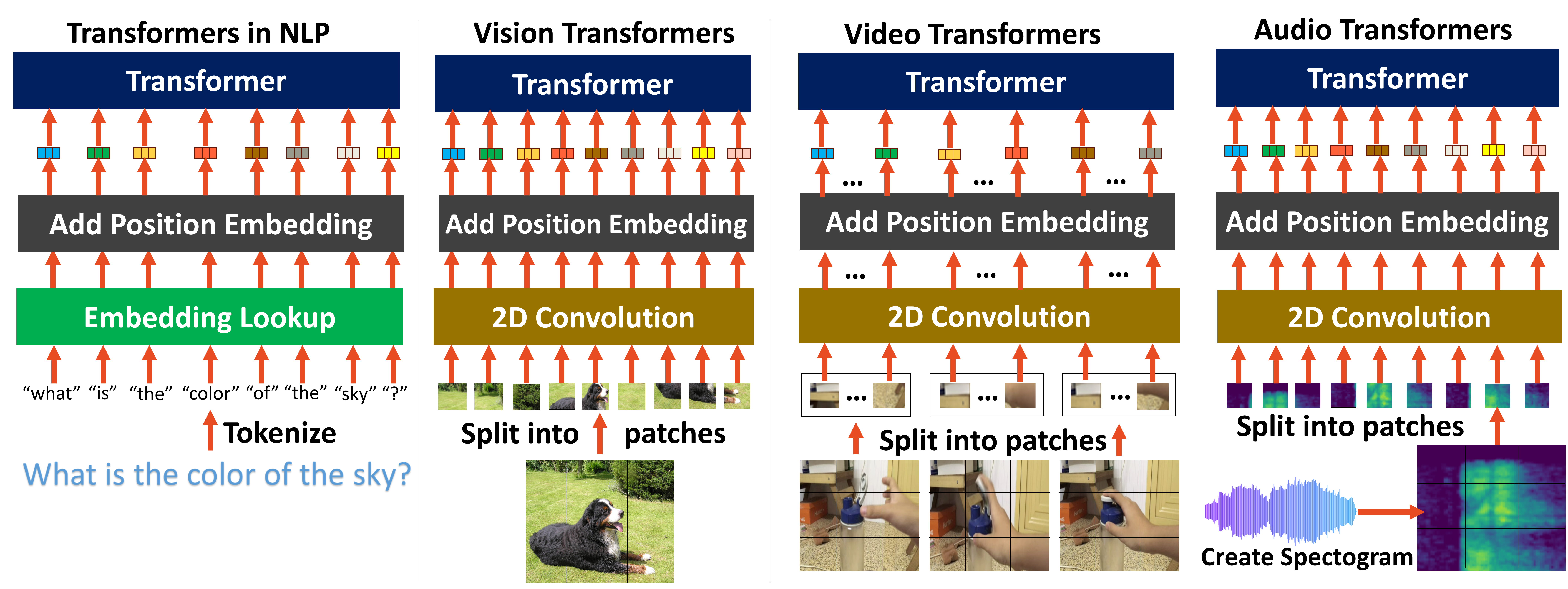}}
\caption{\textbf{Transformers for different modalities.} Similar backbone models are used for all modalities, but different pre-processing steps are necessary to generate embedding vectors from inputs of different modalities.}
\label{fig:overview}
\end{figure*}

Transformers have recently emerged as the state-of-the-art neural network architecture for machine learning tasks involving text, images, video and audio. Remarkably, identical or near-identical Transformer backbones can be used to create high-performance models across a wide range of input modalities \cite{DBLP:conf/naacl/DevlinCLT19, DBLP:conf/iclr/DosovitskiyB0WZ21, DBLP:journals/corr/abs-2303-16058, DBLP:journals/corr/abs-2104-01778}. This is achieved by simply adding a suitable encoding layer which converts the input into a set of embedding vectors, and adds a positional embedding to each vector, thereby outputting a sequence that can be processed by the Transformer (Fig. \ref{fig:overview}). These advantages however come at a cost: Transformers are orders-of-magnitude larger, and hence, more compute-intensive during both training and inference compared to their predecessors such as Convolutional Neural Networks (CNNs).

The number of embedding vectors generated from an input (Fig. \ref{fig:overview}) is the primary factor that determines the computational effort expended by the Transformer. In particular, the computational complexity of self-attention scales quadratically, while all other operations performed by the Transformer scale linearly with number of embedding vectors. We find that the number of embedding vectors generated from an input is directly proportional to the size of the input for all modalities (Fig. \ref{fig:incomp_methods}). For instance, the number of embedding vectors generated from a given text sequence is directly proportional to the number of words in the sequence. On the other hand, the resolution of an image determines the number of embedding vectors generated from it. Similarly, the number of embedding vectors generated from a video depends on the number of frames and the resolution of each frame. Finally, the number of embedding vectors generated from an audio input depends on the number of time steps and the number of frequency banks used to represent the signal at each time step.
In addition to the computational challenges of training Transformers, finding sufficient amounts of data for training them can also be challenging. Transformers are highly susceptible to overfitting for small datasets, as evidenced by the consistent increase in generalization performance with increasing training dataset size \cite{DBLP:journals/corr/abs-2001-08361}. 

We present {\em Input Compression with Positional Consistency} (ICPC), a new augmentation method that simultaneously addresses the efficiency and overfitting challenges of training Transformers. ICPC creates augmented views of each training sample by applying varying levels of compression to it in each epoch. Compression reduces the number of embedding vectors generated from each sample, thereby accelerating training. In contrast, current augmentation methods (such as rotation, translation, CutMix \cite{DBLP:conf/iccv/YunHCOYC19}, mixup \cite{DBLP:conf/iclr/ZhangCDL18}, AugMix~\cite{hendrycks2020augmix}, Manifold Mixup~\cite{verma2019manifold}, etc.) are size-preserving, i.e., they do not change the number of embedding vectors generated per sample.  

ICPC utilizes input compression methods for different modalities that take advantage of their unique characteristics. For text inputs, we propose insignificant word pruning, which reduces the number of embedding vectors by pruning a random subset of less important words from an input sequence in each epoch. For images, we use resolution modulation where images are compressed to random resolutions in each epoch. For videos, we use spatio-temporal resolution modulation to reduce both the number of frames and the resolution of each frame. Finally, for audio signals, we utilize spectogram size modulation, which randomly varies the number of time intervals that the signal is divided into, and the number of frequency banks used to represent the signal in each time interval. 

Since Transformers were initially designed to process text sequences (which can be arbitrarily long), they are inherently capable of processing variable-length inputs. However, we find that na\"ively providing compressed inputs leads to a large drop in accuracy, especially for non-text inputs, due to incorrect encoding of positional information. To overcome this challenge, we propose consistency-aware position selection where the positions associated with each input embedding vector are chosen so as to preserve consistency with the original uncompressed input.

ICPC can also be used to improve the inference efficiency of Transformers through variable-effort inference, wherein samples are first heavily compressed and processed. Easy samples terminate at this stage, while more difficult samples are re-evaluated with lower compression, leading to a net savings in computational effort.

We summarize our main contributions as follows.
\begin{itemize}[leftmargin=*]
\item We propose Input Compression with Positional Consistency (ICPC), a new augmentation method for Transformers that simultaneously improves accuracy and generalization performance.
\item We describe input compression methods for different modalities that reduce the number of embedding vectors generated for each input. 
\item We introduce a consistency-aware position selection method to enable ICPC without any changes to the underlying model architecture. 
\item We demonstrate that training with ICPC consistently improves both accuracy and efficiency over prior methods. We also show that ICPC can be used to improve inference efficiency by modulating computational effort performed based on the difficulty of the input. 
\end{itemize}

%% file: sections/icpc_methods.tex
\section{Data Augmentation through Input Compression}
\label{sec:icpc_methods}

\begin{figure*}[htb]
\centerline{\includegraphics[width=\linewidth]{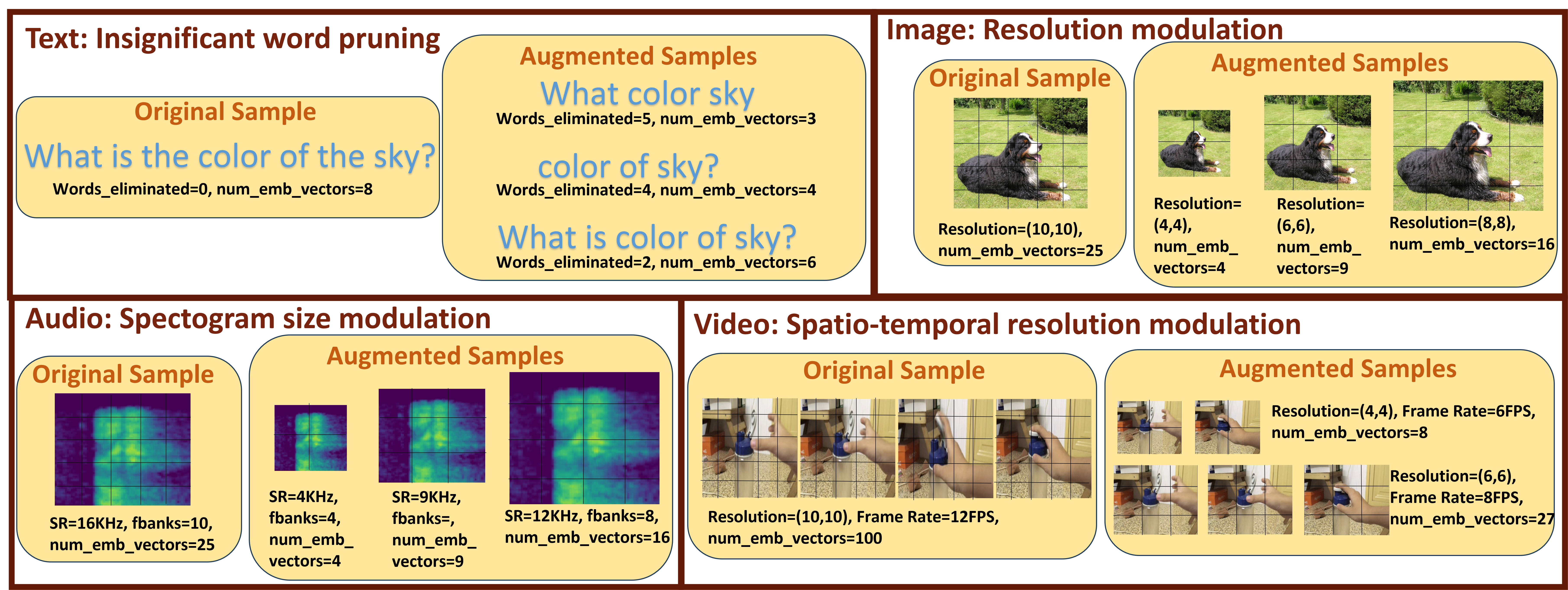}}
\caption{\textbf{Techniques for augmenting data through input compression for different modalities.}}
\label{fig:incomp_methods}
\end{figure*}

The overarching idea behind ICPC is to compress inputs to achieve dual goals --- computational efficiency and data augmentation. While state-of-the-art Transformer models for various modalities use similar architectures as their backbone, they use modality-specific pre-processing steps for encoding the inputs into sequences of embedding vectors. Hence, we propose input compression methods for text, images, video and audio, all of which reduce the number of embedding vectors required to represent a given input (Fig. \ref{fig:incomp_methods}). We describe these methods in turn below.

\noindent\textbf{Text:} Embedding vectors are generated from text sequences by gathering the relevant entries for each word in the sequence from an embedding table that contains entries for all words in the model's vocabulary (Fig. \ref{fig:overview}). Thus, the number of embedding vectors generated from a given sequence is equal to the number of words in the sequence. We propose insignificant word pruning to compress text inputs by removing a random subset of less relevant words in each input sequence in each epoch (Fig. \ref{fig:incomp_methods}). Our procedure for insignificant word pruning is illustrated in Algorithm \ref{alg:algincomp}, lines 1-6. We identify insignificant words in a given sequence using stopword filters. Stopwords are words that do not contribute to the meaning of a sentence, but are required to make them grammatically correct. As a result, pruning stopwords does not affect the labels associated with sequences, thereby enabling augmentation without impacting convergence. We first identify the stopwords in every sequence in the training dataset. Then, the number of stopwords to prune from each sequence in each epoch is determined by selecting a random number between 0 and the total number of stopwords in the sequence. Finally, we randomly select and prune the determined number of stopwords from each sequence. In effect, insignificant word pruning achieves data augmentation by pruning a different subset of stopwords from the sample in every epoch, thereby ensuring that the model does not see the same sequences repeatedly over the course of training. Since pruning stopwords reduces the number of embedding vectors generated from each sequence, insignificant word pruning also improves training efficiency.

\noindent\textbf{Images:} Embedding vectors are generated from images by first splitting them into non-overlapping fixed-size regions called patches, and subsequently extracting an embedding vector from each patch through 2-D convolution. Therefore, the resolution of an image (height, width) determines the number of patches generated, which in turn, determines the embedding vectors generated from the image. We propose resolution modulation for compression-based augmentation of image data (Fig. \ref{fig:incomp_methods}), illustrated in Algorithm \ref{alg:algincomp}, lines 7-11. We start by choosing two random numbers for each batch, with one number indicating the height and the other number indicating the width that all images in the batch will be resized to. Then, all images in the batch are resized to the chosen (height, width) values through downsampling. We restrict ourselves to determining resolution at the batch granularity in order to avoid the padding and ineffectual computations introduced when images in a batch are not of the same size. 

\noindent\textbf{Audio:} Audio signals are represented using spectograms, and embedding vectors are generated by treating the spectogram as a 2-D image and following the method prescribed for images. The number of embedding vectors generated from a given audio signal depends on two factors: the width of the spectogram is determined by the number of time intervals the audio is segmented into (which we refer to as the sampling rate), and the height of the spectogram is determined by the number of frequency banks used to represent the signal. We propose spectogram size modulation for compressing audio inputs (Fig. \ref{fig:incomp_methods}), with the procedure described in Algorithm \ref{alg:algincomp}, lines 20-25. Spectogram size modulation incorporates both sampling rate modulation and filterbank size modulation. We randomly select a sampling rate and number of filterbanks for each batch. Then, each audio sample in the batch is sampled using the selected sampling rate, and converted to a spectogram with the selected number of filter banks at each time step. The sampling rate and number of filterbanks are chosen on a per-batch basis to avoid padding.

\noindent\textbf{Video:} Videos are represented as a series of images (or frames) ordered in time, and embedding vectors are generated by applying the same method prescribed for images to each frame. Consequently, the number of embedding vectors generated from a given video depends on two factors: the number of frames used to represent the video, and the resolution of each frame. Therefore, two forms of compression are possible in videos: (1) spatial compression, which involves reducing the resolution of each frame, and (2) temporal compression, which involves reducing the number of frames. We propose spatio-temporal resolution modulation for augmenting video samples (Fig. \ref{fig:incomp_methods}), with the procedure described in Algorithm \ref{alg:algincomp}, lines 12-19. For each batch in each epoch, we select a random number of frames and spatial resolution (height, width) for each frame. Then, all video samples in the batch are uniformly sampled to generate the selected number of frames, and each frame is subsequently rescaled to the selected (height, width) values. The number of frames and resolution are chosen on a per-batch basis to avoid padding and ineffectual computations. 

\begin{algorithm*}[htbp]
\small
\caption{Data Augmentation through Input Compression for different modalities}
\label{alg:algincomp}
  \DontPrintSemicolon
  
  \SetKwFunction{FMain}{Insignificant word pruning}
  \SetKwProg{Fn}{Function}{:}{}
  \Fn{\FMain{batch}}{
    \For{sequence in batch}
    {
     stopwords = identify\_stopwords(sequence) \\ 
     num\_stopwords\_to\_prune = random(low=0, high=(length(stopwords)-1) \\
     stopwords\_to\_prune = random\_select(stopwords, length=num\_stopwords\_to\_prune) \\
     sequence = sequence - stopwords\_to\_prune \\
    }
  }

    \SetKwFunction{FMain}{Resolution modulation}
  \SetKwProg{Fn}{Function}{:}{}
  \Fn{\FMain{batch, all\_valid\_heights, all\_valid\_widths}}{
    batch\_height = random\_select(all\_valid\_heights, length=1) \\
    batch\_width = random\_select(all\_valid\_widths, length=1) \\
    \For{image in batch}
    {
     image = resize(image, resolution=(batch\_height, batch\_width)) \\ 
    }
  }

    \SetKwFunction{FMain}{Spatio-temporal modulation}
  \SetKwProg{Fn}{Function}{:}{}
  \Fn{\FMain{batch, all\_valid\_heights, all\_valid\_widths, all\_valid\_frame\_rates}}{
    batch\_height = random\_select(all\_valid\_heights, length=1) \\
    batch\_width = random\_select(all\_valid\_widths, length=1) \\
    batch\_frame\_rate = random\_select(all\_valid\_frame\_rates, length=1) \\
    \For{video in batch}
    {
     video = generate\_frames(video, frame\_rate=batch\_frame\_rate) \\
     \For{frame in video}
     {
     frame = resize(frame, resolution=(batch\_height, batch\_width)) \\
     }
    }
  }

    \SetKwFunction{FMain}{Spectogram size modulation}
  \SetKwProg{Fn}{Function}{:}{}
  \Fn{\FMain{batch, all\_valid\_sampling\_rates, all\_valid\_filterbank\_sizes}}{
    batch\_sampling\_rate = random\_select(all\_valid\_frame\_rates, length=1) \\
    batch\_filterbank\_size = random\_select(all\_valid\_filterbank\_sizes, length=1) \\
    \For{audio\_signal in batch}
    {
     sampled\_audio\_signal = sample(audio\_signal, sampling\_rate=batch\_sampling\_rate) \\
     spectogram = create\_spectogram(sampled\_audio\_signal, num\_filterbanks=batch\_filterbank\_size) \\ 
    }
  }
  
\end{algorithm*}

%% file: sections/pos_embed_selection.tex
\section{Consistency-aware position selection: Enabling ICPC in an architecture-agnostic manner}
\label{sec:pos_embed_selection}

Transformers are inherently capable of processing variable-length inputs, i.e., input samples with different numbers of embedding vectors, since they were originally designed to process text inputs that can be arbitrarily long. As a result, inputs presented to the Transformer can have different sizes. In contrast, all inputs presented to CNNs and RNNs are required to be of the same size, since fully-connected layers used in these models require fixed-size inputs. However, we find that position embeddings must be carefully selected to encode inputs whose sizes are smaller than the maximum size supported by the Transformer. In particular, we find that the relative positions of the position embeddings selected to encode compressed inputs must be consistent with the relative positions of vectors generated from the compressed inputs along all dimensions. Consequently, we propose a consistency-aware position selection method that finds the correct subset of position embeddings in the original model for encoding compressed inputs. We describe our position embedding selection methods for different modalities in turn below.

\begin{figure*}[htb]
\centerline{\includegraphics[width=\linewidth]{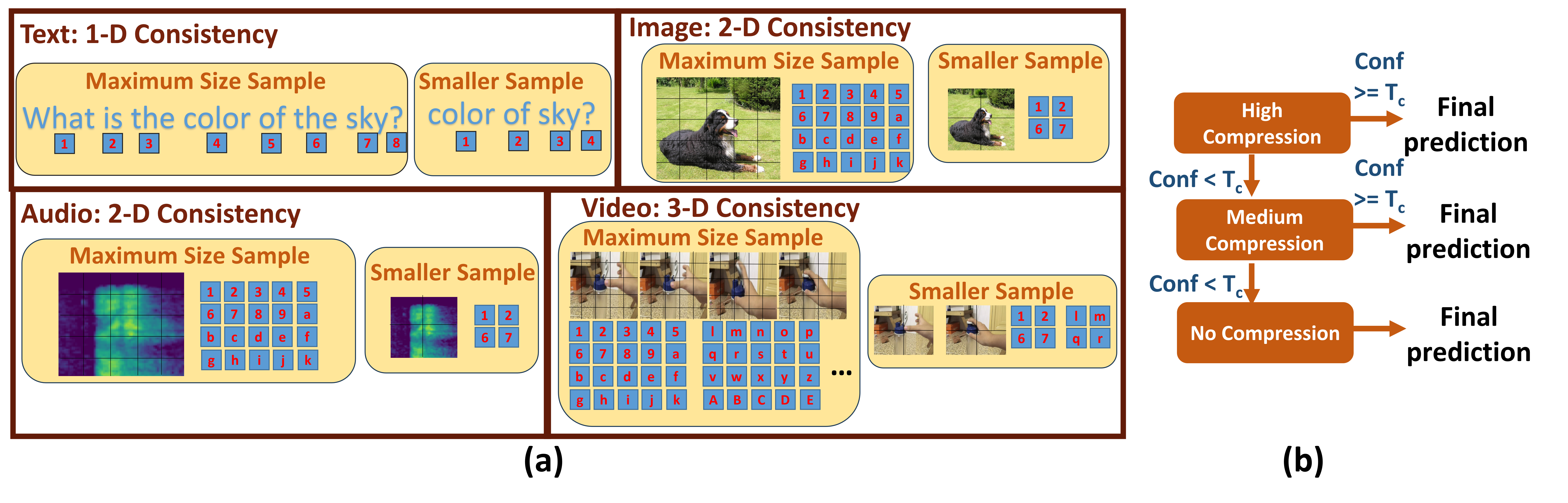}}
\caption{\textbf{(a) Consistency-aware position selection for different modalities.} Blue rectangles represent position embeddings, and letters/numbers represent their positions in the position embedding table. \textbf{(b) Variable-effort inference using ICPC.}}
\label{fig:pos_select}
\end{figure*}

\begin{figure*}[htb]
\centerline{\includegraphics[width=\linewidth]{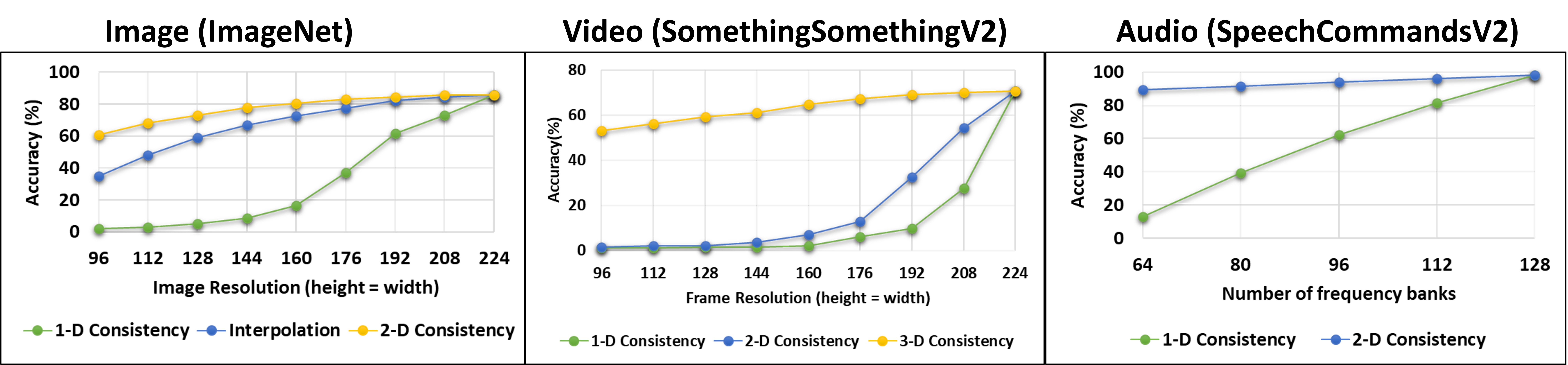}}
\caption{\textbf{Impact of position selection scheme on accuracy when processing compressed inputs.} Results are obtained using fine-tuned models downloaded from the respective repositories. The image model is trained using 224*224 images. The video model is trained using eight 224*224 frames per video. The audio model is trained using a sampling rate of 16KHz and 128 filterbanks. For images, we also compare with the "interpolation" method described in \cite{DBLP:conf/iclr/DosovitskiyB0WZ21} for fine-tuning at a different resolution than the one used for pre-training.}
\label{fig:consistency}
\end{figure*}

\noindent\textbf{Text:} Text inputs are 1-D arrays of words. Since Transformers were designed to process variable-length text inputs, they incorporate a position embedding selection mechanism for inputs that are shorter than the maximum length supported by the Transformer that is designed to maintain 1-D consistency between words in the sequence. For an input sequence of length $n$, 1-D consistency is achieved by selecting the first $n$ entries from the position embedding table (corresponding to the first $n$ positions in a sequence with length equal to the maximum length supported by the Transformer), and encoding the words in the order in which they appear (Fig. \ref{fig:pos_select}). For instance, if three words (A, B, C) appear in that order in a sequence, they are encoded with embeddings corresponding to the following positions: position(B) = 1 + position(A), and position(C) = 1 + position(B).  

\noindent\textbf{Images:} Patches derived from images are arranged into a 1-D stream and fed to the Transformer (Fig. \ref{fig:overview}). Here, we find that simply selecting the first $n$ entries from the position embedding table (as done for text) does not adequately capture the relative positions of patches (Fig. \ref{fig:consistency}). We find that images must be viewed as 2-D grids of patches for accurately selecting position embeddings, since the position of a patch relative to other patches cannot be uniquely determined in 1-D. For instance, the first-$n$ selection method described above cannot encode the fact that two patches are adjacent to each other along the y-axis in the 2-D grid. To address this challenge, we propose a position embedding selection method designed to maintain 2-D consistency between patches in compressed images (Fig. \ref{fig:pos_select}). In particular, if patch A is adjacent to patch B along the x-axis and adjacent to patch C along the y-axis in the 2-D grid, we encode these patches with embeddings corresponding to the following positions -- position(B) = 1 + position(A) and position(C) = width of 2-D grid + position(A) -- thereby encoding adjacency information along both the x- and y-dimensions. 

\noindent\textbf{Audio:} Audio signals are represented using spectograms. Since spectograms are treated as 2-D images during pre-processing, the method described above for encoding images works for encoding patches generated from spectograms also (Fig. \ref{fig:pos_select}, Fig. \ref{fig:consistency}). In particular, if patch A is adjacent to patch B along the time-axis and adjacent to patch C along the frequency-axis in the 2-D grid, we encode these patches with embeddings corresponding to the following positions: position(B) = 1 + position(A) and position(C) = width of 2-D grid + position(A).

\noindent\textbf{Video:} Patches derived from videos are also arranged into a 1-D stream and fed to the Transformer (Fig. \ref{fig:overview}), similar to images. However, we find that the position of each patch relative to other patches can only be accurately captured in 3-D. In particular, encoding compressed videos with a set of 1-D consistent position embeddings (as done with text) only captures the relative positions of patches along the x-axis; adjacency of patches along the y- and time-axes are not captured. 2-D consistent position embeddings (used for encoding images) can capture the relative positions of patches along the x- and y-axes, but not along the time axis (Fig. \ref{fig:consistency}). Consequently, we propose a position embedding selection method designed to maintain 3-D consistency by viewing videos as 3-D grids of patches (Fig. \ref{fig:pos_select}). If patch A is adjacent to patch B along the x-axis, adjacent to patch C along the y-axis and adjacent to patch D along the time-axis in the 3-D grid, our method encodes these patches with embeddings corresponding to the following positions -- position(B) = 1 + position(A), position(C) = width of 2-D grid representing each frame + position(A) and position(D) = (width of 2-D grid * height of 2-D grid representing each frame) + position(A) -- thereby encoding adjacency information along the x- and y-, and time-axes. 

%% file: sections/icpc_inference.tex
\section{Efficient variable-effort inference using ICPC}
When Transformers are deployed for inference, existing methods reshape all input samples to the same shape. As a result, all inputs are represented using the same number of embedding vectors, leading to the same amount of compute time and energy being expended on all samples. However, we observe that many samples can be accurately processed even when they are heavily compressed (represented using only a small number of embedding vectors), and hence, these "easy" samples can be processed at substantially lower computational cost. We find that this is especially true in models trained with ICPC, since training with compressed inputs substantially improves resilience to input compression during inference. Consequently, we propose a variable-effort inference framework that uses ICPC to modulate the computational effort based on the difficulty of each sample (Fig \ref{fig:pos_select}). When a sample is presented during inference, it is first heavily compressed and presented to the Transformer. Only if the Transformer is not confident in predicting the compressed sample ($confidence < T_c$), a less compressed version of the sample is presented to the model. Here, $confidence$ denotes the class probability of the predicted class after softmax and $T_c$ is a hyperparameter that controls the level of confidence required to terminate execution. We describe our variable-effort inference strategies for different modalities in the following subsections.

\noindent\textbf{Images, Video and Audio:} Images are first inferred at low resolution, and are subsequently processed at higher resolution only when necessary, i.e., when the confidence of the Transformer in predicting the low resolution image is less than the confidence threshold. Similarly, videos are first processed using small numbers of frames and low frame resolutions. Higher numbers of frames and resolutions are used only for difficult inputs. Audio signals are initially sampled with low sampling rates and represented using a small number of filterbanks. Both quantities are then progressively increased only for samples that cannot be confidently predicted by the Transformer when compressed. 

\noindent\textbf{Text:} Text sequences are first heavily compressed by pruning all stopwords from each sequence. If the model is not confident in processing the heavily compressed sequence, the amount of compression applied is reduced. One approach to reducing compression is to prune a random subset of stopwords from each sequence, instead of pruning all stopwords. However, we observe that not all stopwords are equally unimportant. For instance, the word "an" is a context-independent stopword, i.e., "an" is irrelevant irrespective of the context it appears in. On the other hand, words such as "beyond" are context-dependent stopwords, i.e., they are irrelevant in most contexts, but are meaningful when the relative positions between certain objects is important for accurately processing the sequence. Based on this observation, we create an ordered list of stopwords based on their relative significance, which we call the Word Importance Hierarchy (WIH). The WIH is created by analyzing the impact of dropping each stopword on the accuracy of a pre-trained model. The stopwords are then arranged in increasing order of accuracy loss incurred by their pruning. Subsequently, different compression levels are created during inference by pruning the first-$n$ stopwords from the WIH from each sequence. In effect, the use of WIH substantially improves the probability of achieving high-confidence predictions when processing compressed inputs compared to random stopword pruning.

%% file: sections/expts_and_results.tex
\section{Experiments and Results}
We implement ICPC in PyTorch, and perform experiments on 4 NVIDIA A40 GPUs, each with 48 GB memory. We use a batch size of 1 during inference, similar to prior works on variable-effort inference \cite{DBLP:conf/icpr/Teerapittayanon16}. We randomly sample 5\% of the training dataset with class balance, and use this as the validation set for determining the confidence threshold ($T_{c}$) for variable-effort inference.  

\noindent\textbf{Experiments on text:} We use the Roberta-Base model \cite{DBLP:journals/corr/abs-1907-11692} along with the stopword list from NLTK \cite{DBLP:conf/acl/Bird06}. We create the WIH by testing a pre-trained Roberta-Base model (downloaded from \cite{DBLP:journals/corr/abs-1910-03771}) on MNLI. During inference, heavy compression is achieved by pruning all words from the stopword list from each sequence. Medium compression is achieved by pruning only those stopwords that lead to a $<=$1\% accuracy drop on the pre-trained model.

\noindent\textbf{Experiments on images:} We use the ViT-Base-224 \cite{DBLP:conf/iclr/DosovitskiyB0WZ21} model for ImageNet, and the ViT-Base-384 model \cite{DBLP:conf/iclr/DosovitskiyB0WZ21} for CIFAR10 and CIFAR100 (both models are pre-trained on ImageNet-21K). During training, the height and width of each image is randomly chosen from [96, 112, 128, ..., 224/384]. Images are resized to (112*112, 176*176) and (192*192, 304*304) for inference with (high, medium) compression on ImageNet and CIFAR, respectively. 

\noindent\textbf{Experiments on video:} We use the UMT-Base-patch16-224 model \cite{DBLP:journals/corr/abs-2303-16058} pre-trained on Kinetics710. During training, the number of frames is randomly chosen from [4, 5, 6, 7, 8], and videos are uniformly sampled to generate the selected number of frames. The height and width of each frame is then randomly chosen from [96, 112, 128, 144, 160, 176, 192, 208, 224]. Videos are represented using (5, 7) frames and frames are resized to (112*112, 176*176) for inference with high and medium compression, respectively. 

\noindent\textbf{Experiments on audio:} We use the AST model \cite{DBLP:journals/corr/abs-2104-01778} pre-trained on ImageNet for SpeechCommandsv2, and the AST model pre-trained on AudioSet for ESC50. During training, the sampling rate is randomly chosen from [8, 9, 10, 11, 12, 13, 14, 15, 16]KHz, and the number of filterbanks is chosen randomly from [65, 75, 85, 95, 105, 115, 125, 128]. Audio signals are sampled at (10, 14)KHz and represented using (75, 105) filterbanks for inference with high and medium compression, respectively. 

\subsection {Primary Results}
We present results of training and inference with ICPC on classification tasks spanning multiple modalities in Table \ref{tab:primary}. For text, we present results on sentiment analysis (SST-2 \cite{wang-etal-2018-glue}) and text categorization (Reuters-21578 \cite{DBLP:conf/iaai/HayesW90}). We present results of image classification on CIFAR-10 \cite{CIFAR10}, CIFAR-100 \cite{CIFAR10} and ImageNet \cite{DBLP:conf/cvpr/DengDSLL009}. We present results on video action recognition using the SomethingSomethingV2 \cite{DBLP:conf/iccv/GoyalKMMWKHFYMH17} and Kinetics400 \cite{DBLP:journals/corr/KayCSZHVVGBNSZ17} datasets, and on speech recognition and environment sound classification using the SpeechCommandsV2 \cite{DBLP:journals/corr/abs-1804-03209} and ESC50 \cite{DBLP:conf/mm/Piczak15} datasets, respectively. We find that using ICPC during both training and inference improves accuracy by up to 1\%, while also accelerating training and inference by up to 2.9$\times$ and 2.6$\times$, respectively.

We observe two complementary sources of accuracy improvement: (1) The additional augmentation from ICPC during training leads to a 0.15\% average accuracy gain across the 9 tasks when all samples are processed without any compression during inference. (2) \textbf{Applying ICPC during inference leads to an additional 0.2\% average accuracy gain}. The accuracy improvement from using ICPC for inference is surprising, since it indicates that for a given sample, the largest possible size is not always optimal. In fact, some inputs are processed more accurately when they are compressed. We hypothesize that this is because the pre-processing steps that generate embedding vectors from inputs can be seen as a form of feature extraction, where each embedding vector represents some feature(s) of the input. When embedding vectors generated from compressed inputs capture the salient features of the input better than the embedding vectors generated from non-compressed inputs, input compression also improves accuracy. During variable-effort inference with ICPC, the confidence of the Transformer in predicting a sample can be viewed as an assessment of the quality of features extracted from the sample. In effect, our method greedily identifies the ideal compression level for each input by progressively reducing compression until sufficiently good features are obtained, thereby simultaneously improving accuracy and efficiency. In fact, we find that $>$75\% of samples have confidence $>$= $T_{c}$ at high compression, and $<$15\% of samples need to be processed with no compression in all studied tasks.

\begin{table}[htbp]
\centering
\caption{\textbf{Results of training and inference with Transformers for different modalities using ICPC.} For the baselines, we follow the exact hyperparameter settings suggested by the authors. ICPC entries are generated by using ICPC during both training and inference. During training, ICPC is used as an additional augmentation method (in addition to the augmenters used in the baseline).}
\resizebox{\linewidth}{!}{%
\begin{tabular}{@{}cccccc@{}}
\hline
\multirow{2}{*}{Modality}  & \multirow{2}{*}{Dataset}  & \multirow{2}{*}{Baseline} & \multirow{2}{*}{ICPC} & Training & Inference \\
          &          &          &      &  Speedup & Speedup \\
\hline
\multirow{2}{*}{Text} & SST-2 & 94.16 & \textbf{94.78} & 1.3$\times$ & 1.7$\times$ \\
 & Reuters & 84.1 & \textbf{84.9} & 1.5$\times$ & 1.6$\times$ \\
\hline

\multirow{3}{*}{Image} & CIFAR-10 & 98.84 & \textbf{99.21} & 2.4$\times$ & 2.2$\times$ \\
 & CIFAR-100 & 92.36 & \textbf{93.31} & 2.3$\times$ & 1.9$\times$ \\
 & ImageNet & 85.84 & \textbf{86.28} & 2.4$\times$ & 1.9$\times$ \\
\hline

\multirow{2}{*}{Video} & SomethingSomethingV2 & 70.76 & \textbf{71.07} & 2.9$\times$ & 2.6$\times$ \\
 & Kinetics400 & 87.42 & \textbf{87.63} & 2.8$\times$ & 2.2$\times$ \\
\hline

\multirow{2}{*}{Audio} & SpeechCommandsV2 & 98.12 & \textbf{98.22} & 1.5$\times$ & 1.3$\times$ \\
 & ESC50 & 95.75 & \textbf{95.89} & 1.6$\times$ & 1.3$\times$ \\
\hline
 
\end{tabular}
}%
\label{tab:primary}
\end{table}

\subsection {Ablation: Evaluation of ICPC as an augmenter}
\begin{figure*}[htbp]
\centerline{\includegraphics[width=\linewidth]{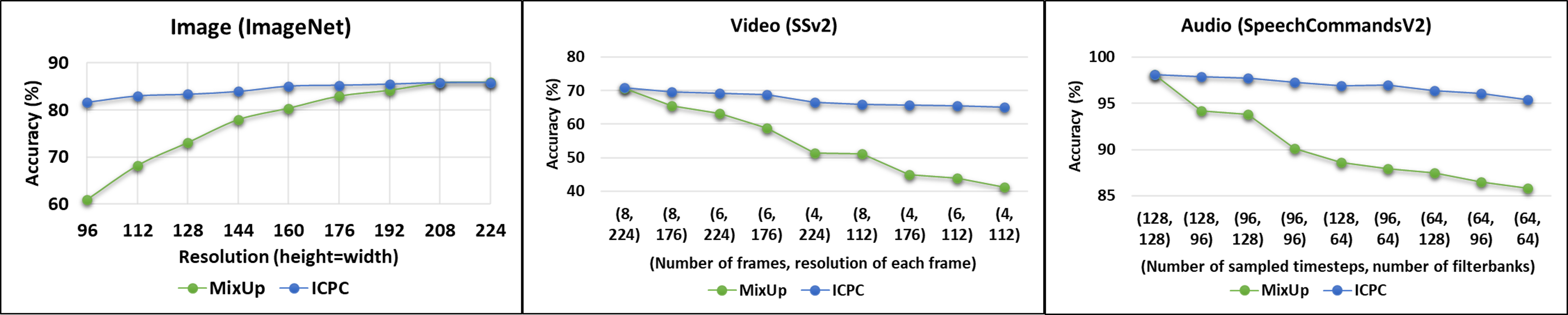}}
\caption{\textbf{Impact of input compression on models trained with and without ICPC.} MixUp is not used when training with ICPC, and vice-versa in this experiment. Our consistency-aware position embedding selection method is used for both cases.}
\label{fig:icpc_varres}
\end{figure*}

We compare ICPC with MixUp \cite{DBLP:conf/iclr/ZhangCDL18}, a popular augmentation strategy that is used for image, video and audio inputs, in Fig. \ref{fig:icpc_varres}. When input compression is not applied during inference, we find that models trained with ICPC are iso-accurate to models trained with MixUp (difference in accuracy is $<$0.5\% for all tasks, with ICPC achieving higher accuracy on 5 out of the 9 studied tasks). However, ICPC simultaneously accelerates training, while MixUp does not improve training efficiency since all composite inputs are resized to fixed shapes. In addition, we find that training with ICPC substantially improves the resilience of models to test-time input compression (Fig. \ref{fig:icpc_varres}). The extent of input compression performed during inference can be tuned to operate at different points in the accuracy-efficiency trade-off space based on user constraints, and ICPC-trained models are significantly more accurate than iso-efficient MixUp-trained models under all levels of compression.

\subsection{Further improving accuracy with Hardware-aware Test-time Augmentation}

\begin{figure}[htbp]
\centerline{\includegraphics[width=0.75\linewidth]{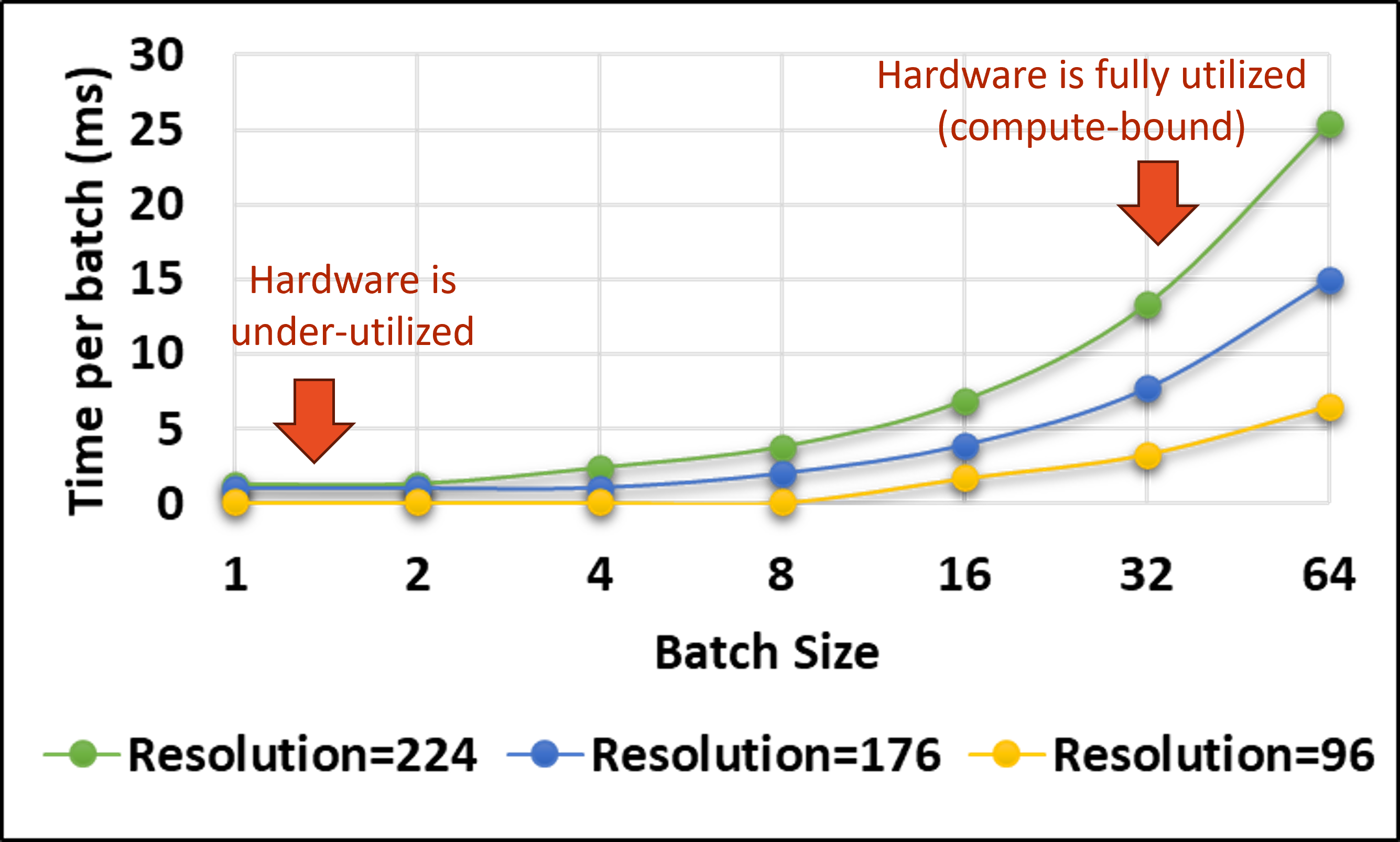}}
\caption{\textbf{Impact of increasing batch size on inference latency for the ViT-Base-224 model on a NVIDIA A40 GPU.}}
\label{fig:comp_bound}
\end{figure}

When a sample is presented during inference, multiple "views" of the sample can be generated using Test-time Augmentation, i.e., by applying the augmentation methods used during training to the sample. Then, predictions on different views of the sample can be combined using an ensembling function (such as averaging, majority voting, etc.) to obtain the final prediction, thereby improving accuracy and robustness. However, the time taken to process each sample increases linearly with the number of augmented views generated from the sample. To address this challenge, we propose Hardware-aware Test-time Augmentation, which takes advantage of hardware under-utilization during inference to enable Test-time Augmentation with minimal increase in latency. In particular, hardware is under-utilized when small batch sizes are used (Fig. \ref{fig:comp_bound}), and increasing the batch size does not increase latency till the batch size is high enough to fully utilize the available compute resources. We term the smallest batch size where the hardware is fully utilized as the "ideal batch size". Latency typically remains constant (or changes very minimally) for all batch sizes less than the ideal batch size. We implement Test-time Augmentation by creating as many views of each sample as possible so that the batch expands to the ideal batch size. 

Our procedure for Hardware-aware Test-time Augmentation is as follows: (1) We use ICPC to augment samples. Since inputs at different compression levels generate different numbers of embedding vectors, padding is used to equalize the lengths of all inputs for batching. Subsequently, attention masks are applied in attention layers to prevent padding vectors from interfering with the processing of valid vectors. (2) The ideal batch size varies with input resolution (Fig. \ref{fig:comp_bound}), with larger ideal batch sizes for smaller inputs. Therefore, there is a trade-off between number of augmented views that can be used for inference at iso-latency, and the maximum resolution of the augmented samples. To find the configuration with the best trade-off, we randomly create K different configurations, with each configuration having a different set of resolutions (number of resolutions in each configuration is equal to the ideal batch size for the maximum resolution in the configuration). For instance, [176, 160, 144, 128] is a valid configuration for ViT-Base on ImageNet, since the ideal batch size is 4 when the resolution is 176*176 (Fig. \ref{fig:comp_bound}). All inputs are evaluated at resolutions of 176, 160, 144 and 128, and the predictions are combined to produce the final prediction when this configuration is used. (3) All K configurations are evaluated on our validation set (5\% of the training set randomly sampled with class balance), and the configuration with the best validation accuracy is evaluated on the test set (Table \ref{tab:hatta}). We find that Test-time Augmentation using ICPC leads to an average accuracy gain of 1.4 absolute points, which is 0.6 absolute points higher than the average accuracy gain from Test-time Augmentation through other augmenters used to train the baseline models. We also find that configurations with lower maximum resolution and higher ideal batch sizes (better for efficiency since the processing time for a batch is primarily dependent on the maximum resolution of samples in the batch) achieve higher average accuracy than configurations with higher resolutions and lower ideal batch sizes.

\begin{table}[htbp]
\centering
\caption{\textbf{Results of Hardware-Aware Test-time Augmentation using ICPC.} We create K=100 configurations, and choose the best one using the validation set. Speedups and accuracy gains (absolute points) are reported over the original (baseline) models.}
\resizebox{0.65\linewidth}{!}{%
\begin{tabular}{@{}ccc@{}}
\hline
Dataset  & Accuracy Gain & Speedup \\

\hline
SST-2 & 1.7 & 1.4$\times$ \\
Reuters & 2.0 & 1.2$\times$ \\
CIFAR-10 & 0.6 & 1.8$\times$ \\
CIFAR-100 & 1.7 & 1.6$\times$ \\
ImageNet & 1.2 & 1.6$\times$ \\
SomethingSomethingV2 & 1.8 & 1.9$\times$\\
Kinetics400 & 1.0 & 2.0$\times$ \\
SpeechCommandsV2 & 0.6 & 1.2$\times$ \\
ESC50 & 1.6 & 1.1$\times$ \\
\hline
 
\end{tabular}
}%
\label{tab:hatta}
\end{table}

%% file: sections/related_work.tex
\section{Related Work}
\label{sec:related_work}

\noindent\textbf{Data augmentation:} Data augmentation is a popular technique for preventing overfitting during training. For text inputs, popular augmentation methods include synonym replacement, shuffling, random insertion and deletion \cite{DBLP:conf/emnlp/WeiZ19}, etc. On the other hand, image datasets are commonly augmented through translation, rotation, noise addition, etc. In addition, techniques such as MixUp \cite{DBLP:conf/iclr/ZhangCDL18}, CutMix \cite{DBLP:conf/iccv/YunHCOYC19} and AugMix \cite{DBLP:conf/iclr/HendrycksMCZGL20} achieve data augmentation by mixing different training samples to create composite inputs. Since videos are represented as sets of images ordered in time, augmentation techniques designed for images have been shown to work well for videos also. Finally, audio datasets are commonly augmented by adding background noise, and by randomly masking out parts of the spectogram \cite{DBLP:conf/interspeech/ParkCZCZCL19, DBLP:journals/corr/abs-2110-05069}. ICPC, which applies varying levels of compression to create augmented views of each sample, is complementary to and can be used in conjunction with the aforementioned augmentation methods. In addition, the vast majority of prior augmentation methods are size-preserving, i.e, the transformations do not change the shape of the input, and hence, they do not have any impact on efficiency.

\noindent\textbf{Transformers for variable-length inputs:} Prior works have proposed modifications to position embeddings in Transformers to enable processing of variable length inputs. SegFormer \cite{DBLP:conf/nips/XieWYAAL21} enables semantic segmentation on variable-resolution inputs through a position-embedding-free model design. NaViT \cite{DBLP:journals/corr/abs-2307-06304} uses fractional embeddings to process images at their native aspect ratios. Patchout \cite{DBLP:journals/corr/abs-2110-05069} uses two different sets of position embeddings -- one capturing time information, and the other capturing frequency information -- for encoding variable-size spectograms. Since these methods require specialized architectures, they are not broadly applicable to all Transformers.

\noindent\textbf{Variable-effort inference:} Variable-effort inference modulates computational effort on a per-sample basis \cite{DBLP:journals/pami/HanHSYWW22} by spending less computational effort in processing easy samples compared to difficult samples. The most popular example is early exit \cite{DBLP:conf/icpr/Teerapittayanon16}, which modulates network depth based on sample difficulty. While early exit modulates model complexity for each sample, ICPC takes a complementary data-centric approach and modulates input sizes. \cite{DBLP:conf/nips/WangHSHH21} varies patch sizes based on input difficulty, but is not applicable to modalities that do not involve patches (such as text).                                                                 

%% file: sections/conclusion.tex
\section{Conclusion}
We proposed Input Compression with Positional Consistency (ICPC), a new data augmentation method that applies varying levels of compression to each sample in every epoch. We introduced a consistency-aware position selection method for encoding compressed inputs. We demonstrated that ICPC improved both generalization performance and training efficiency. We also showed that ICPC can be used to accelerate inference by modulating compression based on input complexity. 